\documentclass[runningheads]{llncs}

\usepackage[T1]{fontenc}

\usepackage{graphicx}

\usepackage{booktabs} 

\usepackage{graphicx}
\usepackage{multirow}
\usepackage{color}
\usepackage{soul}
\usepackage{changes}
\usepackage{kotex}
\usepackage{subcaption}

\usepackage{amsfonts}
\usepackage{amsmath}

\usepackage{algorithm}
\usepackage{algorithmic}

\usepackage{adjustbox}

\usepackage{microtype}

\usepackage{paracol}

\usepackage{eso-pic}
\AddToShipoutPicture*{%
    \AtTextLowerLeft{%
        \makebox[347.5pt][r]{\hspace{\textwidth}\hspace{0cm}\raisebox{0cm}{\textbf{Published in ICPR 2024}}}%
    }%
}

\begin{document}

\title{A Framework for Mining Collectively-Behaving Bots in MMORPGs}

\author{Hyunsoo Kim\thanks{Corresponding author} \and
Jun Hee Kim\thanks{Equal contribution} \and
Jaeman Son$^{\star\star}$ \and 
Jihoon Song \and 
Eunjo Lee$^{\star}$ }

\authorrunning{H. Kim et al.}

\institute{NCSOFT, Republic of Korea\\
\email{\{aitch25,junheekim,jaemanson,songjh7919,gimmesilver\}@ncsoft.com}}

\maketitle              

\newcommand\tc{black}
\newcommand{\hs}[2][]{\st{#1}\color{red}#2 \color{black}}
\newcommand{\rv}[2][]{\st{#1}\color{black}#2 \color{black}}

\begin{abstract}
In MMORPGs (Massively Multiplayer Online Role-Playing Games), abnormal players (bots) using unauthorized automated programs to carry out pre-defined behaviors systematically and repeatedly are commonly observed. Bots usually engage in these activities to gain in-game money, which they eventually trade for real money outside the game. Such abusive activities negatively impact the in-game experiences of legitimate users since bots monopolize specific hunting areas and obtain valuable items. Thus, detecting abnormal players is a significant task for game companies. Motivated by the fact that bots tend to behave collectively with similar in-game trajectories due to the auto-programs, we developed BotTRep, a framework that comprises trajectory representation learning followed by clustering using a completely unlabeled in-game trajectory dataset. Our model aims to learn representations for in-game trajectory sequences so that players with contextually similar trajectories have closer embeddings. Then, by applying DBSCAN to these representations and visualizing the corresponding moving patterns, our framework ultimately assists game masters in identifying and banning bots.

\end{abstract}


\section{Introduction}
\label{sec: Introduction}

In MMORPGs (Massively Multiplayer Online Role-Playing Games), player activities naturally generate diverse patterns, similar to those in the real world. \rv{They can undertake various tasks individually or with others.} \rv{Furthermore, there are groups within the game that carry out activities with malicious intent, as in the real world. The collective behaviors of bots, which exhibit abnormal gaming efficiency due to auto-programs, negatively impact the in-game experiences of regular players. Bots not only monopolize many aspects of the game but also participate in real money trading, which disrupts the in-game economy \cite{huhh2008simple,lee2018no}.

In this study, we introduce a framework for mining collectively-behaving bots, one of the most prevalent forms of abuse in MMORPGs inspired by moving-together patterns in the real world \cite{chen2004marriage,chen2005robust,feng2016survey,li2018deep,tedjopurnomo2021similar,vlachos2002discovering,wang2024deep,zheng2015trajectory}. However, the collectively-behaving groups we aim to identify differ from real-world patterns due to the unique behaviors of bots, such as automatic and sporadic actions needed for purchasing potions, strategic hunting, and returning from dying.} Additionally, teleportation in MMORPGs complicates the detection of these patterns, skewing results that traditional real-world methods might yield. Consequently, our defined collectively-behaving clusters comprise groups of players who not only engage in synchronized activities to optimize farming efficiency \rv{but also display suspicious sporadic behaviors driven by situational demands.}

\rv{Despite the crucial need for fast and accurate bot detection from a service perspective, identifying these bots based on deep learning models is challenging for several reasons:} 1) Real-time labeling of the various trajectories observed in new forms daily is difficult. 2) To establish sufficient evidence, observing whether group movements occur for at least an entire day is necessary, resulting in long sequences. \rv{3) Furthermore, to ensure stable service operation, we must begin monitoring at 9 AM the day after an update to reflect any newly added regions in the game. This means that the training time must be within 9 hours.} 4) False positive detections can bring a loss of trust from users and also cause legal issues (e.g. lawsuit after falsely banning a benign user), and hence an effective methodology for a comprehensive understanding of many users is required. Therefore, we propose a framework to effectively address these industrial challenges, with our contributions as follows:
\begin{itemize}
    \item This framework mines collectively-behaving bots even without labels, proposing a method for trajectory representation learning and DBSCAN \cite{ester1996density}.
    \item The model is designed for efficiency, allowing it to train on long datasets covering an entire day in a shorter time compared to traditional models.
    \item An effective visualization methodology is proposed to quickly double-check if the detected group activity patterns are genuinely collective, contributing to more precise operations.
\end{itemize}
To demonstrate the performance and design validity of our model, we primarily use actual gameplay data from Lineage W\footnote{https://lineagew.plaync.com}, which is an MMORPG released by NCSOFT in November 2021 and is ranked 1 in “Top Grossing Games Worldwide for H1 2022” (Google Play Revenue)\footnote{https://sensortower.com/blog/app-revenue-and-downloads-1h-2022}.

We refer to the proposed model as ``\textbf{BotTRep},'' which stands for a \textbf{T}rajectory \textbf{Rep}resentation model designed to mine \textbf{Bot}s in the game world.

\section{Background}
\label{sec: Background and Basic Idea}

\subsection{\rv{Trajectory data mining for real world tasks}}
\subsubsection{Related works}
\rv{There are various research fields and applications in trajectory data mining \cite{chen2005robust,chen2011discovering,feng2017poi2vec,li2018deep,tedjopurnomo2021similar,vlachos2002discovering,wang2024deep,wang2017region,yan2017itdl,yao2017serm,yin2019gps2vec,zheng2010geolife,zhou2018deepmove}. Even though our research aligns closely with studies such as \cite{li2018deep,tedjopurnomo2021similar,wang2024deep}, we had to design a new mechanism because our research objectives differ from those of trajectory mining in two key aspects.} Firstly, the collectively-behaving bots we aim to detect are not merely groups with similar movement trajectories. While there may be subsets within the collectively-behaving bots that have generally similar trajectories, the bots we need to identify exhibit sporadic behaviors, such as some players replenishing potions in the village or returning to the hunting ground after being killed. Bots have diverse patterns of collective behaviors. Consequently, methods that model representations based on Euclidean space, assuming that two trajectories are similar if they are close in time and space, are fundamentally inappropriate for MMORPGs. Moreover, the coordinate systems in MMORPGs are based on a local coordinate system, making spatial features in Euclidean space incompatible. These are elaborated in the following paragraph. \rv{Secondly, we needed to focus not only on how to extract appropriate trajectory representations but also on how to efficiently train on lengthy sequences. However, studies addressing real-world problems \cite{li2018deep,tedjopurnomo2021similar,wang2024deep} primarily focus on solving issues related to shorter sequences, without considering the challenges posed by longer sequences.}

\vskip -10mm

\subsubsection{Data compatibility issues in two worlds}
\label{subsec: Model compatibility issues in two worlds}

Several issues arise when applying representation learning proposed for real-world data to our task data. The first problem is the existence of teleportation in MMORPGs. Teleportation, a technique that allows for instant movement between two distant spaces, completely overturns our conventional understanding of ``distance''. For instance, players use teleportation to travel from the village to a hunting ground and briefly return to the village to purchase potions during hunting. Thus, while the distance between the village and the hunting ground may not be short, from the perspective of actual behavior in context, they are relatively close. Fig. 1 provides an example of a common trajectory challenge encountered in Lineage W. \rv{In this scenario, entities (b) and (c) are collectively-behaving bots, with entity (c) having died during hunting, revived in the village, and then automatically returned to the hunting grounds.} In contrast, entity (a) represents a player with no affiliation to (b) and (c), whose path coincidentally overlaps with that of (b) for a portion of their movement. In this context, the question arises regarding which entity, (a) or (c), should have embeddings more closely aligned with (b). Models based on spatial features would likely find the embeddings of (a) and (b) more similar due to the physical proximity of their trajectories. However, considering the behavioral context of (b) and (c), their embeddings should be closer. Considering this, we implemented our model by leveraging the contextual relationship between regions instead of Euclidean distance. Thus, in this study, two different spaces frequently visited by the same player are considered ``close'' contextually.

Secondly, the coordinate systems of the two worlds are different. For example, consider two different continents on Earth. The coordinates between the first and the second continents share a coordinate system similar to the concept of a global coordinate system. However, the case is slightly different in MMORPGs. In MMORPGs, for the convenience of game design, some spaces are implemented to have a local coordinate system. For instance, in Linage W, many instance dungeons share the same coordinate range. That is, although each instance dungeon is an independent space, their coordinate ranges overlap, and players in different instance dungeons are recorded as they were in the same space. \rv{This means that it is impossible to use the coordinate features defined on the assumption of spatial proximity as suggested in \cite{li2018deep,tedjopurnomo2021similar,wang2024deep} directly.}

\renewcommand{\thesubfigure}{Fig. \arabic{subfigure}}

\makeatletter
\newcommand{\barefigref}[1]{\arabic{subfigure}} 
\makeatother

\captionsetup[subfigure]{labelformat=simple, labelsep=colon}

\begin{figure}[htp]
\vskip -6mm
\centering
\subfloat[Spatially, entities (a) and (b) are in close proximity; however, if the overall context between (b) and (c) is similar, then the embeddings of (b) and (c) should be closer.\label{fig: similar_traj}] {
  \includegraphics[height=3.1cm]{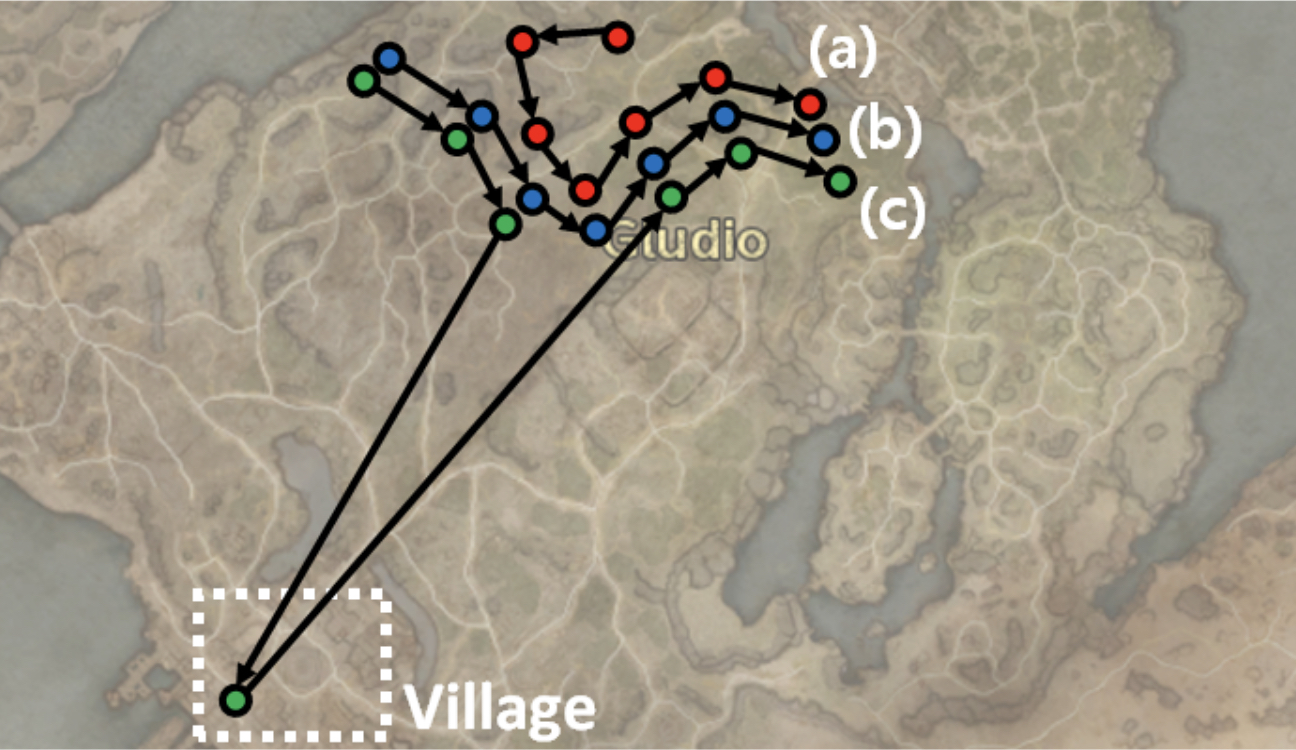}}
  \hfill
\subfloat[This figure represents the outcome of binning applied to coordinate values in Lineage W. We bin coordinate values into zones and cells of different sizes and embed them. \label{fig: binning}]{
  \includegraphics[height=3.1cm]{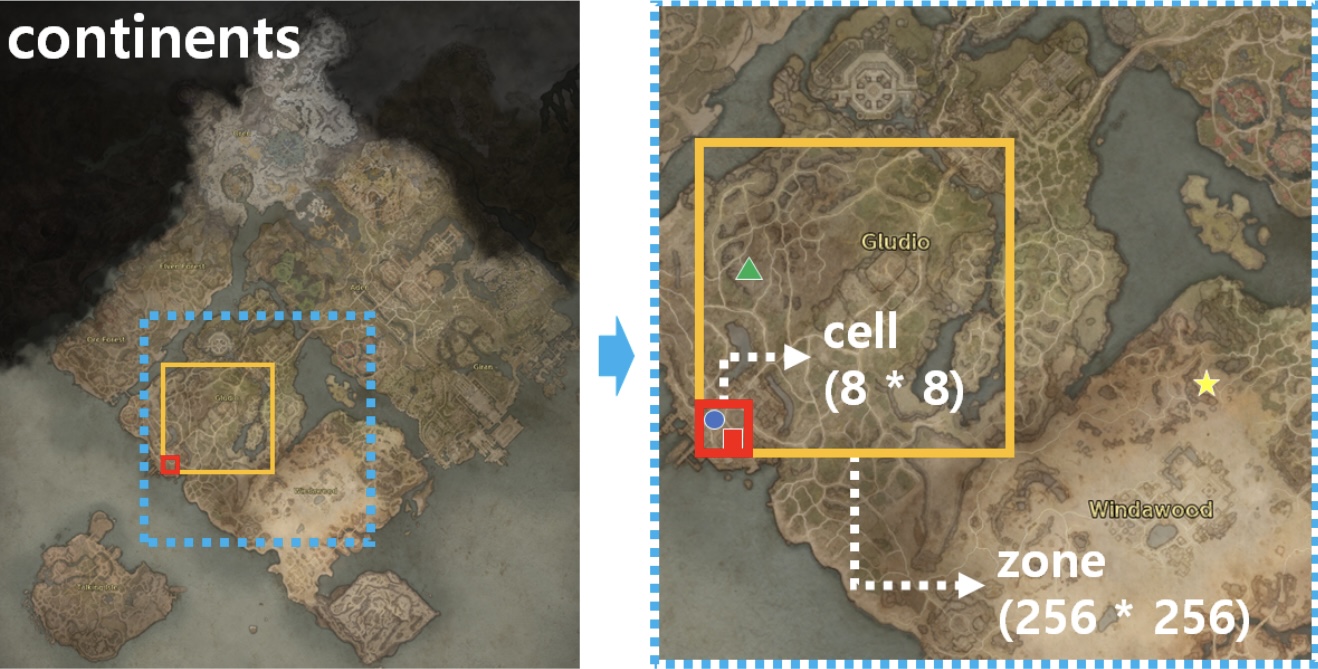}}
\vskip -8mm
\end{figure}

\subsection{\rv{Bot detection and trajectory mining for MMORPGs}}

\subsubsection{Related works}
The mentioned works utilize various features for classifying abnormal players, such as logs from player's status (portrait), events, quests, tradings, mouse clicks, and trajectory \cite{chen2009game,lee2018no,pao2010game,qi2022gnn,su2022trajectory,tao2019mvan,zhao2022t}. Amongst the works, this study aimed at trajectory mining for bot detection in MMORPGs \cite{chen2009game,pao2010game,su2022trajectory,zhao2022t}. These works focused on detecting bots using their in-game trajectories with various approaches. For instance, \cite{chen2009game,pao2010game} designed models to classify the bots from benign players inspired by repetitive and regular observations generated by bots with various in-game features, such as lingering, smoothness, detour route, and turn angle. On the other hand, \cite{su2022trajectory,zhao2022t} utilized deep-learning-based approaches to classify the players with suspicious trajectory patterns.

Notably, compared to the other related works, our framework puts emphasis on providing explainable materials for game masters. For example, authors in \cite{chen2009game,pao2010game,su2022trajectory,zhao2022t} focused only on constructing accurate models to classify bots and benign players using their trajectories and other features. Consequently, their features were preprocessed into an unexplainable form, which is suitable for deep learning models but challenging for human beings to understand. Particularly, the authors in \cite{su2022trajectory,zhao2022t} dynamically set the size of areas corresponding to each trajectory token based on the frequency of visits to each region and utilized preprocessed time information with diverse event types, such as finger touches or mouse clicks. These features showed respectable performances in downstream tasks but were complicated for model explanation.

\subsubsection{The necessity of explainability in industrial applications}
The current trend focuses on automation using deep learning, and some companies aim to exclusively use deep learning models for automatically sanctioning abnormal users. Most studies related to gaming bots have concentrated on how to detect the bots with high accuracy. Their research is significant; however, in practical industries, there is a persistent question not just about whether a specific player group is bots, but why they are bots. When there is clear evidence, game masters can go ahead and ban the detected users with lower legal risk. Therefore, in the context of bot detection, the interpretability of model results and the ability to provide evidence are crucial. Especially in real world scenarios, if game masters mistakenly ban legitimate users, they can face legal repercussions, and the company may set unfavorable precedents. Moreover, even if game masters have properly sanctioned bots, failing to provide evidence for the reasons they were classified as bots when the bots' owners file a lawsuit also sets a bad precedent. For this reason, we propose a framework that does not fully automate the process but instead aids game masters in more efficiently detecting abnormal players.

\section{Proposed Approach}
\label{sec: Proposed Approach}

\subsection{\rv{Preliminary}}
\label{subsec: Basic idea}

\subsubsection{Defining areas}
\label{subsubsec: Defining areas in game worlds}

In this section, we introduce three geographical terms to distinguish areas in the game world: continent, zone, and cell. A continent is the largest area category, classified by whether players can move on foot or need to use a portal or teleport. For example, islands and instance dungeons are treated as separate continents. We define zones by dividing continents into multiple areas, each sized at 256 by 256 coordinates, as shown in Fig. 2. Similarly, zones contain several cells, each sized at 8 by 8 coordinates. In Lineage W, players can move about 256 coordinates per minute. The main continent measures 2048 by 2048 coordinates, and there are over 100 continents, including islands and instance dungeons, each sized between 256 by 256 and 512 by 512. \rv{When logging location coordinates in the game, the unique continent ID where the player was located at each timestamp and the detailed coordinates within that continent are recorded.}

Our model is trained by zone and cell tokens, which provide spatial information. Zones offer abstract representations for larger areas, while cells provide specific details for smaller areas. Training the model with only zone tokens reduces the out-of-vocabulary issue but hampers its ability to distinguish between different trajectories. Conversely, training with only cell tokens allows discrimination between trajectories but leads to the out-of-vocabulary problem and unstable convergence in model training.

\rv{We determined the appropriate criteria empirically during model design. We recommend setting the width and height of a zone to be half the size of an instance dungeon to prevent tokens from being overly abstracted. The width and height of a cell should be approximately the range of a ranged character, such as a mage or archer. This ensures that the movement of ranged characters, as well as groups of both ranged and melee characters, can be detected more precisely.}

\subsubsection{The definition of collectively-behaving bots}
\label{subsubsec: The definition of collectively-behaving bots}

As outlined earlier, our goal is to detect collectively behaving bots as accurately and extensively as possible. The term ``collectively-behaving bots'' refers to groups of 4 or more players exhibiting evidence of group activities throughout their session. Our proposed framework is designed to identify such behavior clusters.

\subsubsection{Contrastive model}
\rv{We propose a model for detecting collectively-behaving bots using a contrastive approach. The model learns to make the representations of similar trajectory inputs closer together while pushing the representation vectors of dissimilar trajectory inputs further apart. The reasons for choosing a contrastive model are clear. First, contrastive models yield more robust trajectory representations \cite{wang2020understanding}. Second, the task requires faster training times. Existing models for real-world problems \cite{li2018deep,tedjopurnomo2021similar,wang2024deep} typically use autoencoder structures, which lack a direct procedure for distinguishing similar and dissimilar sequence pairs. Additionally, autoencoders are heavy and slow due to their encoder-decoder structure. While complex autoencoder structures are suitable for tasks that require understanding precise relationships between tokens, our task prioritizes extracting appropriate representations of trajectory sequences over token relationships. Thus, we propose a lightweight contrastive model that ensures faster convergence and superior performance specifically for this task.}

\subsection{Data preparation}
\label{subsec: Data preparation}

\subsubsection{Traning dataset}
The game logs we used in this work consist of coordinates and timestamp logs sampled at one-minute intervals. This means that if a user played the game for an entire day, we would sample 1,440 logs—one for each minute of the day.\footnote{24 hours a day is 1440 minutes.} As mentioned earlier, the game logs we aim to train on are significantly lengthy. Consequently, instead of feeding the entire log sequence into the model for training, we have preprocessed the structure of the input data to ensure the model can effectively learn the relationships between contextually close cells appearing in each sequence. To achieve this, we extract a data point in the training dataset based on the following rules:
\begin{enumerate}
    \item \rv{Collect data for all locations where players have visited on a daily basis.}
    \item \rv{Utilize the collected coordinates logs to generate tokens for zones and cells.} 
    \item Construct a sequence using the generated zone and cell tokens, ensuring that neighboring cells within the sequence are not identical, to include various local information within a sequence.
    \item \rv{Split the sequence into multiple data points, each with a length of 32.}
    \item \rv{Reorganize the preprocessed data with a length of 32 into triplet formats in two modes: 1) odd-even split mode, which uses odd and even indexes, and 2) half split mode, which uses the first half and second half indexes. In each mode, the anchor $(\mathcal{A})$, positive $(\mathcal{P})$, and negative samples $(\mathcal{N})$ will have a length of 16 each.}
    \item Finally, masking is applied to the preprocessed anchor sequences from the previous step. The masking occurs with a probability of $r$ (where $r \in \{0.2, 0.3\}$) for the sequence tokens.
\end{enumerate}

\rv{Specifically, we preprocessed the training sequences to a length of 16 to ensure the model effectively learns the differences between token sequences in each positive and negative sample. When constructing anchor, positive, and negative samples based on two split modes, a longer input sequence would contain too much regional information, causing the token types in positive and negative samples to become similar. Additionally, the similarity of the trajectories between the anchor and positive samples would decrease, especially in the half-split mode. Consequently, the model would struggle to clearly learn the differences between positive and negative samples.}

To express the process mathematically, we first define ($L^{p_1}$) in equation (1) as the raw coordinate location sequence of a specific player $p_1$, \rv{where $p_1 \in \mathbf{P}$, and $\mathbf{P}$ is the entire set of players. Here, $\{p_1, .., p_{N} \} = \mathbf{P}$, and $N$ is the number of players. An element $l_{i}$ (where $l_{i} \in L^{p_1}$) is in the form of ($x, y$) coordinate pair with continent ID ($c$): ($l^{(x)}_{i},\ l^{(y)}_{i},\ l^{(c)}_{i}$). The continent ID addresses the design issue of the local coordinate system, where two players located in different spaces could be recorded as being at the same coordinates.

\vskip -4mm
\begin{align}
L^{p_1} =& \left \{(l^{(x)}_{1}, l^{(y)}_{1}, l^{(c)}_{1}), (l^{(x)}_{2}, l^{(y)}_{2}, l^{(c)}_{2}), .., (l^{(x)}_{|L^{p_1}|}, l^{(y)}_{|L^{p_1}|}, l^{(c)}_{|L^{p_1}|}) \right \} = \left \{l_1, l_2, .., l_{|L^{p_1}|} \right \}
\label{eq: dataset1}
\end{align}
} 
\vskip -4mm

Now, we generate a sequence for representation learning by selecting some elements in $L^{p_1}$ corresponding to the conditions in equation (2). Equation (2) is included to ensure that diverse cell tokens are incorporated into a single sequence. Then, we define $S^{p_1}$ in (3) as entire lengths of preprocessed binned sequences of $p_1$. The elements of $L^{'p_1}$ are binned into bins named zone and cell by \rv{applying two functions: $\mathbf{z}(l^{'}_i)=(\left \lfloor l^{'(x)}_{i} / 256 \right \rfloor, \left \lfloor l^{'(y)}_{i} / 256 \right \rfloor, l^{'(c)}_{i})$, and $\mathbf{c}(l^{'}_i)=(\left \lfloor l^{'(x)}_{i} / 8 \right \rfloor, \left \lfloor l^{'(y)}_{i} / 8 \right \rfloor, l^{'(c)}_{i})$.}

\vskip -4mm

\rv{
\begin{align}
L^{'p_1} =& \left \{l_i \ |\ \mathbf{c}(l_{(i-1)}) \neq \mathbf{c}(l_{i}) \ \text{or}\ i=1,\ \text{for}\ i \in \{1, 2, .., |L^{p_1}|\} \right \} \\ 
S^{p_1} =& \left \{(\mathbf{z}(l^{'}_{i}), \mathbf{c}(l^{'}_{i})) \ |\ l^{'}_i \in L^{'p_1}, \ i = 1, 2, .., |L^{'p_1}| \right \} 
\label{eq: dataset2}
\end{align}
where $L^{p_1}$, $L^{'p_1}$, and $S^{p_1}$ are ordered set.}

Here, we rewrite elements of ${S}^{p_1}$ as \rv{${S}^{p_1} = \{(\mathbf{z}(l^{'}_1), \mathbf{c}(l^{'}_1)), (\mathbf{z}(l^{'}_2), \mathbf{c}(l^{'}_2)), .., \allowbreak (\mathbf{z}(l^{'}_{|L^{'p_1}|}), \mathbf{c}(l^{'}_{|L^{'p_1}|}))\} = ({s}_{1}, .., {s}_{|L^{'p_1}|})$} for readability. \rv{Afterward, we split the preprocessed $S^{p_1}$ into $j$ subsequences, each with a length of 32 in equation (4). 

\vskip -4mm

\begin{align}
S^{p_1}_j =& (s_{(j-1) \cdot 32 + 1}, s_{(j-1) \cdot 32 + 2}, \ldots, s_{j \cdot 32})\text{ where }j\text{ is an integer, } 1 \leq j \leq \left \lfloor \frac{|L^{'p_1}|}{32} \right \rfloor
\end{align}
}
\vskip -4mm

\rv{Through this process, we preprocessed the entire daily coordinate sequence of a specific user into sequences of length 32 ($S^{p_1}_j$). This process is then repeatedly applied to all users $\{p_1, .., p_N\}$. The result can be represented as $D^{prep} = \{S^{p_1}_{1}, .., S^{p_1}_{\left \lfloor {|L^{'p_1}|}/{32} \right \rfloor}, .., S^{p_N}_{1}, .., S^{p_N}_{\left \lfloor {|L^{'p_N}|}/{32} \right \rfloor}\}$ = $\{d^{prep}_1, .., d^{prep}_M\}$ where $M$ is the number of preprocessed data point. Afterward, we reorganize the elements $d^{prep}_k$ (where $d^{prep}_k \in D^{prep},\ 1 \leq k \leq M$) into triplet formats ($\mathcal{A}$, $\mathcal{P}$, and $\mathcal{N}$) in two modes as follows:
\begin{paracol}{2}
\begin{leftcolumn}
\subsubsection*{Odd-even split}
\begin{align} 
{\mathcal{A}}^{(o)}_{k}=&\left \{{s}_{i} \in {{msk}_{r}^{\delta}}({d}^{prep}_{k}) | i=1, 3, .., 31 \right \} \nonumber \\ 
{\mathcal{P}}^{(o)}_{k}=& \left \{{s}_{i} \in {d}^{prep}_{k}\ |\ i=2, 4, .., 32\right \} \nonumber \\
{\mathcal{N}}^{(o)}_{k}=&\ \mathcal{P}^{(o)}_{{\substack{k^{'} \\ k \neq k^{'}}}} \text{ where } k^{'} \sim {U}(1, M) \nonumber
\end{align}
\end{leftcolumn}
\begin{rightcolumn}
\subsubsection*{Half split}
\begin{align} 
{\mathcal{A}}^{(h)}_{k} =& \left \{{s}_{i} \in {{msk}_{r}^{\delta}}({d}^{prep}_{k}) | i=1, .., 16\right \} \nonumber \\ 
{\mathcal{P}}^{(h)}_{k} =& \left \{{s}_{i} \in {d}^{prep}_{k}\ |\ i=17, .., 32\right \} \nonumber \\ 
{\mathcal{N}}^{(h)}_{k} =&\ \mathcal{P}^{(h)}_{{\substack{k^{'} \\ k \neq k^{'}}}} \text{ where } k^{'} \sim {U}(1, M) 
\label{eq: mode2}
\end{align}
\end{rightcolumn}
\end{paracol} 

\vskip -4mm

\noindent where $U(1, M)$ is a uniform distribution. That is, when constructing the negative sample for index $k$, we composed the data by assigning the positive sample from a different sample (where $k \neq k^{'}$) out of the total $M$ data points. Next, ${msk}_{r}^{\delta}(\cdot)$ is a function that applies masking to input tokens with a probability of $r$, where $r \in \{0.2, 0.3\}$. The information about which token indices have been masked is recorded in a set $\delta$.

We repeat this process for entire data points ($M$) to create the training dataset ($D^{train}$). That is, $D^{train} = \{({\mathcal{A}^{(o)}_{1}}, {\mathcal{P}^{(o)}_{1}}, {\mathcal{N}^{(o)}_{1}}), .., ({\mathcal{A}^{(o)}_{\left \lfloor M/2 \right \rfloor}}, {\mathcal{P}^{(o)}_{\left \lfloor M/2 \right \rfloor}}, {\mathcal{N}^{(o)}_{\left \lfloor M/2 \right \rfloor}}), .., \allowbreak ({\mathcal{A}^{(h)}_{\left \lfloor M/2 \right \rfloor + 1}}, {\mathcal{P}^{(h)}_{\left \lfloor M/2 \right \rfloor + 1}}, {\mathcal{N}^{(h)}_{\left \lfloor M/2 \right \rfloor + 1}}), .., ({\mathcal{A}^{(h)}_{M}}, {\mathcal{P}^{(h)}_{M}}, {\mathcal{N}^{(h)}_{M}}) \} = \left \{ {d}^{train}_{1}, .., {d}^{train}_{M} \right \}$. 
}

\vskip -12mm

\subsubsection{Dataset for downstream tasks}
\rv{Preprocessing the dataset for downstream tasks is much simpler compared to generating the training data. For downstream tasks, we utilize the raw trajectory sequence ${L}^{p1}$ right after applying binning functions: $\mathcal{T}^{p_1} = \left \{ (\mathbf{z}({l}_{i}), \mathbf{c}({l}_{i})) \ | \ l_i \in L^{p_1},i = 1, .., |L^{p_1}| \right \}$. This process is repeated for all $N$ players, and we notate this dataset as $D^{traj} = \left \{\mathcal{T}^{p_1}, .., \mathcal{T}^{p_N} \right \}$.}

\subsection{Task-specific representation model}
BotTRep utilizes a contrastive structure based on the Transformer architecture \cite{kenton2019bert,vaswani2017attention}. It is trained jointly using the triplet margin loss \cite{schroff2015facenet} and cross-entropy loss for two tasks: contrastive learning and masked cell prediction, as shown in Fig. \ref{fig: contrastive learning model}. The following subsections provide a detailed explanation of the design strategy and the two tasks.

\subsubsection{Transformer-based contrastive learning task}

\setcounter{figure}{2}

\begin{figure}[h!]
\vskip -10mm
\centering
\includegraphics[height=3.1cm]{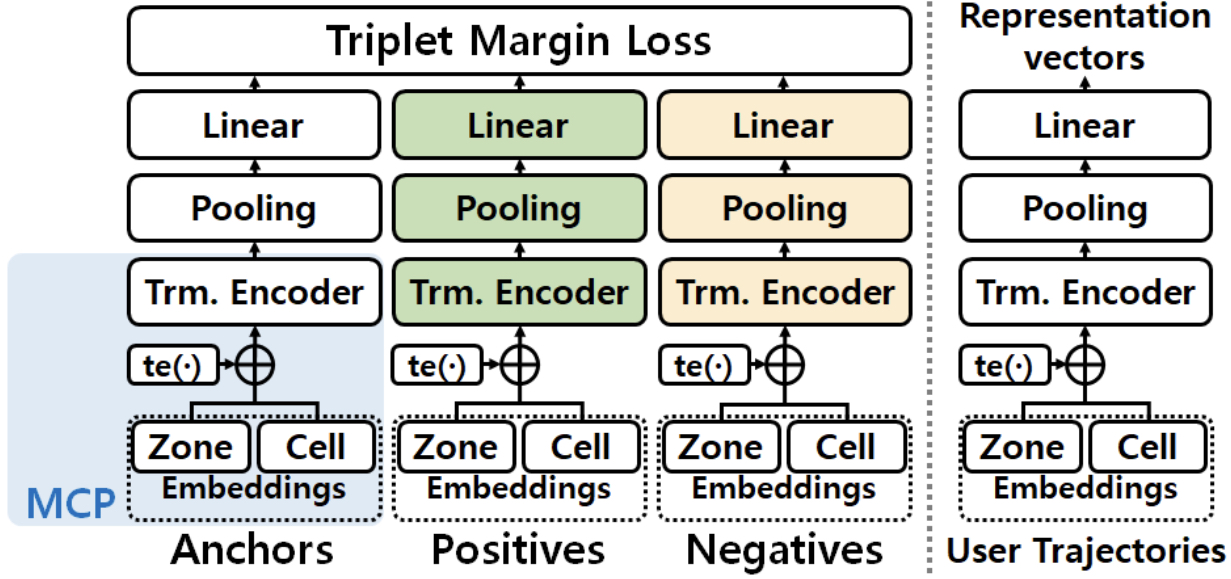} 
\hfill
\includegraphics[height=3.1cm]{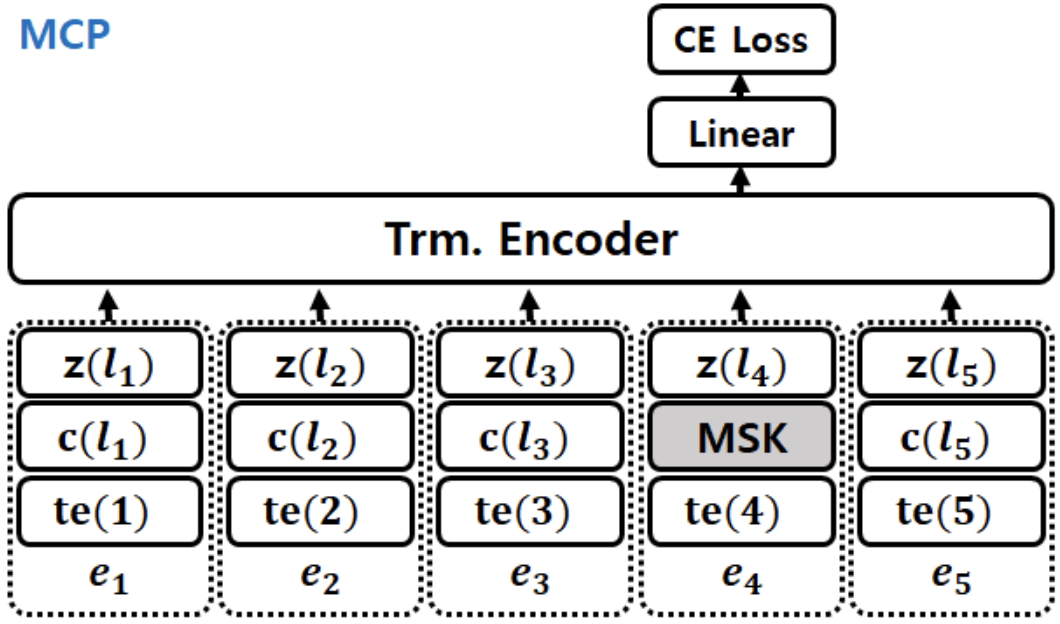} 

\vskip -10pt
\caption{The left side of the figure depicts our Transformer-based model for the contrastive learning task (left) and its representation extractor (right), respectively. The right side of the figure shows how the training for the MCP task is performed.}
\label{fig: contrastive learning model} 
\vskip -4mm
\end{figure}

In this part, we begin by explaining the process of contrastive learning with a data point of zone and cell sequences, \rv{such as $\mathcal{A}_i$ from $(\mathcal{A}_i, \mathcal{P}_i, \mathcal{N}_i) = {d}^{train}_{i}$ where (${d}^{train}_{i} \in D^{train}$}), as described in subsection \ref{subsec: Data preparation}. Embedding modules, $\mathbf{e_z}(\cdot)$ and $\mathbf{e_c}(\cdot)$, map each zone and cell token to $d$-dimensional vectors, and ${e}_{j} \in \mathbb{R}^{d}$ where $d \in \{256, 512\}$. The embedding matrix associated with a sequence, \rv{$\mathcal{A}_{i}=(s_1, .., s_j, .., s_{16})$, is initialized as ${{E}_{i}} = ({e}_{1}, .., {e}_{j} .., {e}_{16})$ where ${e}_{j} = \mathbf{e_z}({s_j}) + \mathbf{e_c}({s_j}) + {\mathbf{te}}{(t+j)}$} where $\mathbf{te}(\cdot)$ is a function for timestamp encoding and $t$ is a random index between 1 and 1424, corresponding to minute indexes in a day. The timestamp encoding has a similar concept as the positional encoding \cite{vaswani2017attention}. The timestamp encoding is also a mapper generated by: $\mathbf{te}{(j, 2m)} = sin({j/10000}^{2m/d})$ and $\mathbf{te}{(j, 2m+1)} = cos({j/10000}^{2m/d})$. Here, we limit the last random index to 1424 because the length of our input sequence is 16, and the input value for $\mathbf{te}(\cdot)$ must not exceed 1440 because it indicates 1440 minutes a day. The reason that we input randomly generated timestamp values is to make the model train from diverse input for each epoch. We set the independent random timestamp for the anchor and negative sample, and for the positive sample, we set the dependent timestamp values from the anchor's timestamp. For example, if we define a uniform random function as \rv{$t \sim {U}(a, b)$, and $t_\mathcal{A}$, $t_\mathcal{P}$, and $t_\mathcal{N}$ as randomly generated timestamps of the first index, we set $t_\mathcal{A} \sim {U}(1, 1424)$,  $t_\mathcal{P} \sim t_\mathcal{A} + {U}(-16, 16)$, and $t_\mathcal{N} \sim {U}(1, 1424)$, respectively. That is, the elements of $\mathcal{A}_i$ are finally initialized as follows: ${e}_{j} = \mathbf{e_z}({s_j}) + \mathbf{e_c}({s_j}) + {\mathbf{te}}{(t_{\mathcal{A}}+j)}$, and the same applies to $\mathcal{P}_i$ and $\mathcal{N}_i$.}

\rv{Now, ${E}_{i}^{\mathcal{A}}$, ${E}_{i}^{\mathcal{P}}$, and ${E}_{i}^{\mathcal{N}}$ are embedding matrices of anchor ($\mathcal{A}$), positive ($\mathcal{P}$), and negative sequences ($\mathcal{N}$), respectively.} In training, we modified the input by changing the ratio of anchor to positive samples to 2:1, allowing the model to learn from a wider variety of sequence combinations.

\rv{Then, embedding matrices (${E}_{i}^{\mathcal{A}}$, ${E}_{i}^{\mathcal{P}}$,  ${E}_{i}^{\mathcal{N}}$) are input to the Transformer encoder layer. For convenience, all these processes are denoted as follows: $F^{\mathcal{A}}_{i} = \mathbf{trm}(E^{\mathcal{A}}_{i})$ where $\mathbf{trm}(\cdot)$ is a Transformer encoder block, and $F^{\mathcal{A}}_{i}$ represents the token-unit output of the anchor sequence produced by the model. Here, the lengths of ${E}^{\mathcal{A}}_{i}$ and ${F}^{\mathcal{A}}_{i}$ are both 16. We set the dimensionality of the inner-layer ($d_{inner}$) to 1024-2048, depending on the embedding dimension. The encoder block is composed of a stack of 8 identical layers in this work. Next, the pooling layer calculates average pooling from the output of the Transformer encoder block, ${F}^{\mathcal{A}}_{i}$, then, the linear layer receives the pooled vector and calculates the final output, ${\mathcal{R}^{\mathcal{A}}_{i}} = Linear(Pooling({F}^\mathcal{A}_{i}))$ where $Linear(x) = x{W}^{T} + b$. ${\mathcal{R}^{\mathcal{A}}_{i}}$ is an example of the output representation of anchor; the positive and negative samples' outputs, ${\mathcal{R}^{\mathcal{P}}_{i}}$ and ${\mathcal{R}^{\mathcal{N}}_{i}}$, are calculated in the same way.} The loss of our proposed model is obtained by the below function, named triplet margin loss (\ref{eq: loss}). This function minimizes the distance between an anchor and a positive sample and maximizes the distance between an anchor and a negative sample. 
\rv{
\begin{align}
\setcounter{equation}{5}
\mathcal{L}_1(\mathcal{R}_i^{\mathcal{A}}, \mathcal{R}_i^{\mathcal{P}}, \mathcal{R}_i^{\mathcal{N}}) = \left[ \left\| \mathcal{R}_i^{\mathcal{A}} - \mathcal{R}_i^{\mathcal{P}} \right\|_2^2 - \left\| \mathcal{R}_i^{\mathcal{A}} - \mathcal{R}_i^{\mathcal{N}} \right\|_2^2 + \beta \right]_+
\label{eq: loss}
\end{align}
}
\vskip -10mm

\subsubsection{Masked cell prediction (MCP) task}
In addition to the contrastive learning task, our model incorporates the masked language model (MLM) task proposed in BERT \cite{kenton2019bert} to refine the learning of cell tokens. However, we named it the ``masked cell prediction'' (MCP) task in our model because the task is no longer related to language models. To apply this, masking is performed on the sequence data before model training, and this masking is only applied to the anchor sequences, as shown in equation (5). In MCP, the problem involves predicting what the token was before being masked in the anchor sequence that has been masked during the preprocessing process.

\rv{Specifically, we apply the $Linear(\cdot)$ function to $F^{\mathcal{A}}_i$, returned by $\mathbf{trm}(\cdot)$, for training. Since the length of the sequences inputted into our model is 16, for convenience, we denote this as $F^{\mathcal{A}}_i = (f^{\mathcal{A}}_1, .., f^{\mathcal{A}}_{16})$.} In the MCP task, among these 16 extracted results, $Linear(\cdot)$ is applied to the tokens that had been masked, and then the loss between the predicted results and the actual answers is calculated using Cross Entropy Loss. The process can be formalized as follows:
\begin{align}
\hat{y}_j = Linear(f^{\mathcal{A}}_j) \quad \text{for each } f^{\mathcal{A}}_j \in F^{\mathcal{A}}_i, \text{ where } j \in \delta
\label{eq: mcp1}
\end{align}
where $\delta$ is the set we recorded masked indexes, and $\hat{y}_j$ is the predicted output for the $j$-th token in the sequence. The objective is to minimize the loss between the predicted output $\hat{y}_j$ and the true label $y_j$ for the tokens that were originally masked. The loss is calculated using Cross Entropy Loss: 
\begin{align}
\rv{\mathcal{L}_{2}({y}_{j}, {\hat{y}}_{j}) = - \sum_{k=1}^{C} y_{j,k} \log(\hat{y}_{j,k})}
\label{eq: mcp2}
\end{align}
\rv{where $y_j$ is the true label for the $j$-th token, $C$ is the number of classes (vocabularies), and $k$ is the index for each class, ranging from 1 to $C$.} The summation is performed over all tokens in the sequence that were masked.
Ultimately, training is conducted by summing the losses calculated from the top two tasks and then performing backpropagation:  $\mathcal{L}  = \mathcal{L}_{1} + \mathcal{L}_{2}$.

\subsection{Extract representation vectors}
\rv{Now, we can extract representations from user trajectories using our trained model.} We utilize the model that extracts user representations in Fig. \ref{fig: contrastive learning model}. The entire procedure is almost the same as the model train step, but the observed timestamp values are added to each token instead of randomly generated values. We utilize timestamp encoding designed on a minute basis, as mentioned previously. Afterward, we extract the daily trajectory of each player from the games, and each data point is denoted as $\mathcal{T}^{p_i}$, as mentioned before. The embedding matrix associated with \rv{$\mathcal{T}^{p_i}$ is also initialized as ${E}^{\mathcal{T}_{p_i}}_{i} = ({e}_{1}, .., {e}_{t})$ where ${e}_{t} = \mathbf{e_z}({\mathbf{z}({l}_{t})}) + \mathbf{e_c}({\mathbf{c}({l}_{t})}) + {\mathbf{te}}{(t)}$ where $t$ is timestamp index from 1 to 1440, stands for the timestamp of $t^{th}$ location that the player has visited.} Then, the embedding matrix ${E}_{i}$ is input to the trained model, and the model extracts representations. To summarize, the simplified structure of our suggestion is $\mathcal{T}^{'p_i} = \mathbf{model}({E}^{\mathcal{T}_{p_i}}_{i})$ where $\mathbf{model(\cdot)}$ is the proposed model, and the representations of the trajectories are notated in this way: ${D^{rep}} = \left \{\mathcal{T}^{'p_1}, \mathcal{T}^{'p_2}, .., \mathcal{T}^{'p_N} \right \}$.

\subsection{Clustering collectively-behaving groups}

Once we obtain representation vectors for the trajectories, we cluster them so that players that have similar in-game trajectories and, hence, similar representations get grouped together. In particular, we use the DBSCAN \cite{ester1996density}. An interesting property of DBSCAN is that not all data points get assigned to a cluster. That is, the algorithm classifies points that do not belong to any group as noise. Such points generally correspond to benign players because they have peculiar trajectories resulting from diverse preferences in play styles and are typically not included in specific clusters. Hence, we decided to define all the clustered groups (i.e. DBSCAN did not classify as noise) as collectively-behaving groups. However, in order to utilize DBSCAN appropriately, optimization of parameters, $min\_samples$ and $eps\ (\varepsilon)$, should be preceded. Here, we select $min\_samples$ as 4 because we target suspicious groups with more than 4 players relying on our industrial requirement. Afterward, we control $\varepsilon$ by referring to the distances of representation vectors of 4 closest neighbors from each data point, inspired by \cite{schubert2017dbscan}. Notably, we chose not to use a fixed $\varepsilon$ value for DBSCAN. Instead, we adopted the methodology from \cite{schubert2017dbscan} that selects $\varepsilon$ values based on the density of representation vectors. This is because the vectors we cluster are derived from the deep learning model. Even when using the same data for training, the inherent randomness in the learning process can result in representation vectors in hidden space having different scales of distance and density among specific data points. In other words, using a fixed $\varepsilon$ value led to issues in maintaining consistent quality when detecting collectively-behaving bots based on the model’s output vectors.

\begin{figure}[h!]
\centering
\includegraphics[width=\linewidth]{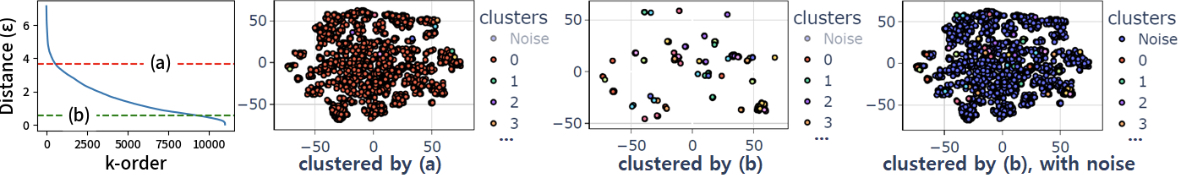} 
\vskip -10pt
\caption{The first image shows a comparison between the $\varepsilon$ selection criterion proposed by \cite{schubert2017dbscan} (a) and our optimized criterion (b). The thrid image illustrates clustering results using criterion (b), visualized with t-SNE \cite{van2008visualizing} for dimension reduction, excluding noise clusters.} 
\label{fig: clustering}   
\vskip -5mm
\end{figure}

Fig. \ref{fig: clustering} explains how we choose the $\varepsilon$ value in the clustering process, and the clusters that result from it. Firstly, the leftmost image shows the criterion for selecting $\varepsilon$ as proposed in \cite{schubert2017dbscan} (a), and the criterion we used in this work (b). \cite{schubert2017dbscan} suggests drawing a plot of distance and then using the distance at the elbow point of the curve as the $\varepsilon$ value. However, applying the value from (a) to our data led to the phenomenon of clustering together trajectories that are relatively dissimilar, as shown in the second plot in Fig. \ref{fig: clustering}. Consequently, we searched for a new criterion (b) suitable for our data and are utilizing this value as the $\varepsilon$ for DBSCAN. Precisely, we extract the distances between each data point and their 4 nearest data points. Afterward, we select a distance value of 0.05-0.20 quantile ($q$) from the extracted distances by K-NN. For example, $\varepsilon = quantile_{0.05}(dist)$ where $quantile_{q}(\cdot)$ is a quantile function that returns $q$ quantile from input data, and $dist = 4-NN_{dist}(D^{rep})$ where $4-NN_{dist}(\cdot)$ is a $k-NN$ algorithm where $k=4$ and returns distances between the 4 closest neighbors from each data point. 

\vskip -4mm

\vskip -10mm

\section{Experiments}
\label{sec: Experiments}

\subsection{Dataset : Lineage W}
We train and evaluate our proposed model with two different datasets: 1) a preprocessed dataset for \rv{model training, and 2) a real-world gaming trajectory dataset for the downstream task.} The first dataset includes 778,656 samples of preprocessed trajectories collected for 8 days on July 1st-8th, 2023. When training, we performed parallel computing on 8 NVIDIA A40 GPUs. We have configured the model to be trained for at least 70 epochs and terminated based on early stopping criteria with patience of 8 epochs. The training time was approximately 6-12 minutes per epoch, depending on the parameters. The second dataset contains 26,136 player trajectories collected for 7 days on July 9th-15th, 2023. \rv{
Tables 1 and 2 present the experimental results for the second dataset.}

\subsection{Evaluation methods}

\subsubsection{Contextual similarity} 

In this work, we evaluate the similarity in representation between two pairs: collectively-behaving pairs and random pairs, which we call positive and negative pairs, respectively. In this step, we prepared the experimental environment by collecting daily trajectory sequences from all players in a game world and extracting their representations. Subsequently, we generated positive and negative groups based on their representations using DBSCAN. Then, we excluded noise-labeled data determined by the DBSCAN algorithm. We then selected positive and negative pairs based on the cluster labels in this way: \rv{
\begin{align}
pos &= \left \{(\mathcal{T}^{p_i}, \mathcal{T}^{p_j}) \ | \ i = 1, .., N,\ j \in {\xi}(\mathcal{T}^{'p_i}) \right \} \\
neg &= \left \{(\mathcal{T}^{p_i}, \mathcal{T}^{p_j}) \ | \ i = {1, .., N}, \ j \notin {\xi}(\mathcal{T}^{'p_i}) \right \}
\label{eq: cs}
\end{align}
where ${\xi}(\cdot)$ is a function that returns the closest index from neighboring players within the cluster containing each input player.
}

After composing experimental data pairs in the above way, we measure the contextual similarity between their trajectories using a metric named time-aware Jaccard similarity. In this study, we design time-aware Jaccard similarity to check whether the pair of trajectories have similarities over time. \rv{We calculate Jaccard similarity \cite{jaccard1912distribution,jadon2023comprehensive} by a 30-minute subset of each trajectory using 1440 range of minute indexes ($t$).} This metric returns the overall similarity by averaging every 30-minute subsets between two trajectories with 15-minute shifts. \rv{The similarity scores from this metric can be interpreted as a measure of how contextually close two different users were in terms of their locations at approximately the same moments. Trajectory pairs with similar travel routes exhibit high similarity values.} Contextual similarity is calculated at the cell level and is defined as follows:
\begin{align} 
\frac{1}{\left | T \right |}\sum J_t(\mathcal{T}^{p_i},\mathcal{T}^{p_j})&\ \text{where}\ t \in \{1, 16, 31, .., 1411\} = T \nonumber \\ 
\text{and}\ J_t(\mathcal{T}^{p_i}, \mathcal{T}^{p_j})& = {{|\{{tr}_{t}^{p_i}, .., {tr}_{t+29}^{p_i}\} \cap \{{tr}_{t}^{p_j}, .., {tr}_{t+29}^{p_j}\}|}\over{|\{{tr}_{t}^{p_i}, .., {tr}_{t+29}^{p_i}\} \cup \{{tr}_{t}^{p_j}, .., {tr}_{t+29}^{p_j}\}|}}
\label{eq: jaccard}
\end{align}

\subsubsection{Access information homogeneity}
The another metric is designed to verify whether there is indeed the same abuser behind the clusters mined as collectively-behaving groups. In addition to user trajectory data, our dataset records access information data for each user, such as the IP and device-sharing network mentioned in \cite{tao2019mvan}. This metric aims to determine whether their access information actually belongs to the same person. In other words, if the access information of players mined as collectively-behaving groups is identical, it signifies that they are indeed real collectively-behaving bots controlled by the same user. The specific calculation method is as follows: $\frac{1}{\left | C \right |}\sum_{c_{id} \in C} acc\_info(c_{id})$, where $C$ represents all clusters excluding a noise cluster, that is, the entire collectively-behaving groups we have mined. $c_{id}$ denotes each collectively-behaving group. $acc\_info(\cdot)$ is a function that takes a collectively-behaving group as input and returns the number of different access information points possessed by players in that cluster. For instance, if there are 4 players within a certain cluster and their access information is interconnected, 1 is returned. However, if, upon checking the access information of the 4 players, it is found that 3 of their access information points are interconnected, but one is different, then they form 2 groups, and thus 2 is returned. \rv{That is, bots controlled by the same owner have identical access information, resulting in low access information homogeneity values. In contrast, legitimate users who operate one avatar at a time have different access information, resulting in high access information homogeneity values.}

\subsection{Ablation study}

This section summarizes the ablation study results for the proposed method. We experimented by altering the model's key parameters. Additionally, we highlight the benefits of incorporating both zone and cell tokens and the Masked Cell Prediction (MCP) technique, assessing their impact on model performance.

To validate this, we varied parameters and documented the outcomes in experiment type (a). We compared models trained on cell inputs alone versus those trained on both zone and cell inputs, and examined the effects of incorporating MCP, with results under experiment type (b). In experiment type (c), we adjusted the clustering parameter $q$ to observe its impact on our model and DBSCAN's performance. As $q$ increases, contextual similarity and homogeneity of access information decrease due to less homogeneous clusters. A larger $q$ adopts a more lenient criterion for detecting suspicious players, while a smaller $q$ is preferred for higher precision.

\subsubsection{Experimental results}

Our proposed setting for BotTRep is shown at the top of Table \ref{tab: ablation} ($\dagger$). This model showed the best performance from the perspective of contextual similarity when we conducted additional experiments adjusting $d\_model$, $d\_hid$, $\beta$, and MCP ratio based on this model; the model generally exhibited higher performances in experiment type (a). 
In experiment type (b), we conducted experiments by removing zone embedding and MCP one at a time from $\dagger$, and a slight decrease in performance was observed in both cases regarding contextual similarity. Lastly, the results from experiment type (c) showed that as clustering $\varepsilon$ quantile ($q$) increased, contextual similarity and access information all decreased. As $q$ increases, one can observe which clusters are included in the suspicious clusters. For example, Fig. \ref{fig: heatmap} on the far right shows a sample of players included in the noise cluster. Their trajectories generally do not display a pattern; however, there are exceptions, such as case (a) in Fig. 5. If we were to set the DBSCAN’s $\varepsilon$ value higher, groups similar to (a) might be included in the collectively-behaving group. \rv{To provide additional context for the metrics, the detected bots exemplified in (\url{https://youtu.be/bsFXvFBVYak}) show an average time-aware Jaccard similarity of around 0.3 across all pairs and an access information homogeneity of 1.0.}

\begin{table}[]
\vskip -8mm
\centering
\caption{The results of the ablation study are shown below, with the best performance highlighted in bold. The dataset contains 26,136 data points.}

\begin{tabular}{ccccccccccc}
\hline
\multirow{2}{*}{\begin{tabular}[c]{@{}c@{}}Exp\\ types\end{tabular}} & \multicolumn{5}{c}{Model params}                    & clustering                 & \multirow{2}{*}{\begin{tabular}[c]{@{}c@{}}Detecting \\ count \end{tabular}} & \multicolumn{2}{c}{Contextual}    & \multirow{2}{*}{\begin{tabular}[c]{@{}c@{}}Acc \\ info \end{tabular}} \\ \cline{2-7} \cline{9-10}
                                                                     & $d\_model$ & $d\_hid$ & $\beta$ & zone  & MCP & $\varepsilon$ quant. ($q$) &                                                                            & $pos$             & $neg$             &                           \\ \hline
$\dagger$                                                            & 256        & 1024     & 0.5     & True  & 0.2       & 0.05                       & 928                                                                        & \textbf{0.3625} & 0.0005          & \textbf{1.0079}           \\ \hline
\multirow{3}{*}{(a)}                                                 & 256        & 1024     & 0.5     & True  & 0.3       & 0.05                       & 921                                                                        & 0.3596          & 0.0005          & 1.0444                    \\
                                                                     & 256        & 1024     & 1.0     & True  & 0.2       & 0.05                       & 976                                                               & 0.3552          & \textbf{0.0003} & \textbf{1.0079}           \\
                                                                     & 512        & 2048     & 0.5     & True  & 0.2       & 0.05                       & 963                                                                        & 0.3560          & \textbf{0.0003} & 1.0697                    \\ \hline
\multirow{3}{*}{(b)}                                                 & 256        & 1024     & 0.5     & True  & 0.0       & 0.05                       & 918                                                                        & 0.3208          & 0.0006          & 1.0787                    \\
                                                                     & 256        & 1024     & 0.5     & False & 0.2       & 0.05                       & 917                                                                        & 0.3472          & 0.0005          & 1.0551                    \\
                                                                     & 256        & 1024     & 0.5     & False & 0.0       & 0.05                       & 930                                                                        & 0.3183          & 0.0006          & 1.1307                    \\ \hline
\multirow{3}{*}{(c)}                                                 & 256        & 1024     & 0.5     & True  & 0.2       & 0.10                       & 1932                                                                       & 0.3203          & 0.0006          & 1.0787                    \\
                                                                     & 256        & 1024     & 0.5     & True  & 0.2       & 0.15                       & 3031                                                                       & 0.2931          & 0.0006          & 1.1094                    \\
                                                                     & 256        & 1024     & 0.5     & True  & 0.2       & 0.20               & \textbf{4109}                                                                       & 0.2732          & 0.0009          & 1.1363                    \\ \hline
\end{tabular}
\label{tab: ablation}
\vskip -8mm
\end{table}
\vskip -8mm

\subsection{Baseline models}

In this study, \rv{we deliberated on selecting the most appropriate baseline model due to the lack of precedent representation models applied to tasks similar to ours.} To compare and validate the performance of our model, we prepared three types of baseline models, as described below. These models employ the encoder-decoder structure most commonly used for sequence data representation, \rv{with internal layers implemented in three variants using Bi-GRU, Bi-LSTM and Transformer blocks.} \rv{We compared our model to autoencoder-based models because they are commonly used for extracting sequence representations. Previous trajectory representation models proposed for GPS data using proximity features in the real world \cite{li2018deep,tedjopurnomo2021similar,wang2024deep} have also used autoencoders. For the dataset, we trained and inferred directly using the data for downstream tasks ($D^{traj}$) without additional preprocessing steps, such as checking if adjacent cells are identical. This decision was made because such preprocessing significantly slowed the model's convergence, leading to poor performance within the time constraints of our requirements. 

} 

When preparing these models, we set $d\_model$ to 256 to enable a comparison with our model under similar specifications. \rv{When extracting representations, we applied mean pooling after the input data passed through the encoder block because it achieved the best performance compared to CLS and max pooling.}

\subsubsection{Experimental results}

Table \ref{tab: baseline} describes the outcomes of the baseline experiments. \rv{Results from these experiments revealed that the Transformer significantly outperforms Bi-GRU and Bi-LSTM.} The Transformer's results in a contextual similarity of 0.29, slightly lower than our model, while the homogeneity of access information stands at 1.79, indicating a somewhat higher figure compared to our proposal. This suggests the inclusion of benign players within the collectively-behaving groups. Transformer could not properly distinguish the two trajectories active in the same area but at different times. From the perspectives of contextual similarity and homogeneity of access information, BotTRep, with our proposed setting ($\dagger$) was found to have the highest performance. This indicates that the access information of all players within a cluster is related, signifying our model has effectively detected the bot groups we aimed to identify with high accuracy. \rv{Furthermore, BotTRep completed training in about 8 hours and 30 minutes, achieving over 70 epochs, while the other two models failed to surpass its performance even after over 24 hours of training.}

\begin{table}[]
\setlength{\tabcolsep}{5.5pt} 
\centering
\vskip -8mm 
\caption{BotTRep showed superior performance in contextual similarity and access information. The downstream dataset contains 26,136 data points.}
\begin{tabular}{ccccccc}
\hline
\multirow{2}{*}{Models} & \multirow{2}{*}{\begin{tabular}[c]{@{}c@{}}Detecting \\ count \end{tabular}} & \multicolumn{2}{c}{Contextual}    & \multirow{2}{*}{Acc info} & \multirow{2}{*}{\begin{tabular}[c]{@{}c@{}}Training\\ time\end{tabular}} & \multirow{2}{*}{\begin{tabular}[c]{@{}c@{}}Minutes\\ (per epoch)\end{tabular}} \\ \cline{3-4}
                        &                                                                                                 & $pos$             & $neg$             &                           &                                                                          &                                                                                \\ \hline
Bi-GRU          & 926                                                                                             & 0.2157          & 0.0006          & 1.8886                    & 24+ hours                                                                & 20-21                                                                          \\
Bi-LSTM        & \textbf{1,064}                                                                                  & 0.1716          & 0.0006          & 3.0315                    & 24+ hours                                                                & 20-21                                                                          \\
Transformer     & 970                                                                                             & 0.2921          & \textbf{0.0002} & 1.7855                    & 24+ hours                                                                & 25-27                                                                          \\
BotTRep ($\dagger$)        & 928                                                                                             & \textbf{0.3625} & 0.0005          & \textbf{1.0079}           & \textbf{8.5 hours}                                                       & 6-7                                                                           \\ \hline
\end{tabular}
\label{tab: baseline}
\vskip -8mm
\end{table}

\subsection{Trajectory visualization}

This work is designed to surveil collectively-behaving groups, supposed to be bots, throughout the continents using their trajectory data and visualize how much their trajectories are similar to each other. The heatmap, designed to show similar colors when players exist contextually close to each other, is used to visualize whether the suspicious players appear together. The heatmap's $x$-axis represents timestamps, and the $y$-axis represents player indices. The color, in RGB format, indicates the player's location at a specific timestamp. That is, we construct a 3-dimensional vector that somehow represents spatial information, and then visualize it as RGB coloring. Specifically, we generate colormaps as ($continent\_lv$, $x$, $y$), where $continent\_lv$ reflects average player levels in each continent. This addition represents semantic relationships between continents. Regarding game geography, hunting grounds suitable for each level are located in an adjacency to guide players in growing their avatars with less confusion. Note that if the player was not logged in the game at a particular timestamp, we color the corresponding spot as white.

\begin{figure}[h!]
\vskip -6mm
\centering
\includegraphics[width=\linewidth]{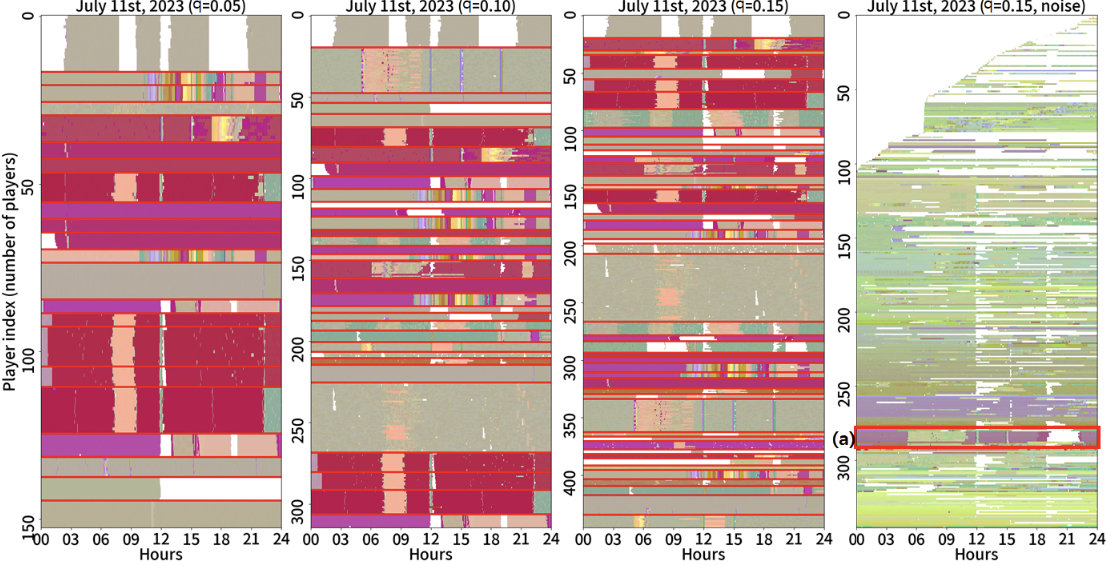} 
\vskip -4mm
\caption{This image shows player locations over time based on clustering results. The $x$-axis represents time in minutes, the $y$-axis indicates individual players, and colors correspond to players' location at each time point. Red lines in the heatmap separate clusters, with similar colors denoting proximity of player locations.} 
\label{fig: heatmap}   
\vskip -8mm
\end{figure}

Finally, we present a visualization of player location over time in Fig. \ref{fig: heatmap}. The right-most plot corresponds to the points such that DBSCAN labeled as noise: i.e. those that we consider benign users. We can see that these users tend to have their own distinct trajectories, indicating that they are indeed not collectively-behaving bots. On the other hand, the three plots on the left show the results for trajectories where DBSCAN assigned a cluster. Each red horizontal line indicates the separation of clusters. Recall that players located closer to each other will exhibit similar colors in the heatmap. Clearly, we can see that players in the same cluster tend to have similar sequences of colors throughout the entire timeline ($x$-axis). That is, they collectively move across multiple areas or collectively log in/log out, both simultaneously and in order. This pattern is a typical characteristic observed in collectively-behaving bots that we have targeted: this arises due to multiple avatars being controlled simultaneously by automated programs connected to the same network environment. Thus, our framework that consists of trajectory embedding and clustering identifies suspicious clusters where the corresponding players move collectively.

\section{Conclusion}

We proposed a novel framework that uses a trajectory representation model trained jointly on contrastive learning and masked cell prediction tasks, so that similar contextual in-game movements obtain closer representations. Then, we used DBSCAN to identify collectively-behaving groups and introduced a visualization method that explains their in-game trajectories. Our framework meets the industrial needs for clear explainability and can assist game masters by providing clustered users who are suspicious to be collectively-behaving bots.


\end{document}